\documentclass{article}

\usepackage{PRIMEarxiv}

\usepackage[utf8]{inputenc} 
\usepackage[T1]{fontenc}    
\usepackage{hyperref}       
\usepackage{url}            
\usepackage{booktabs}       
\usepackage{amsfonts}       
\usepackage{nicefrac}       
\usepackage{microtype}      
\usepackage{lipsum}
\usepackage{fancyhdr}       
\usepackage{graphicx}       
\graphicspath{{media/}}     

\title{Beyond Color and Lines: Zero-Shot Style-Specific Image Variations \\ with Coordinated Semantics
\thanks{\textit{\underline{Citation}}: 
\textbf{Authors. Title. Pages.... DOI:000000/11111.}} 
}

\author{
  Jinghao Hu, Yuhe Zhang, GuoHua Geng, Liuyuxin Yang, JiaRui Yan, Jingtao Cheng, YaDong Zhang, Kang Li  \\
  School of Information Science\&Technology \\
  Northwest University \\
  Xi'an, Shaanxi Province, China\\
  \texttt{2017118127@stumail.nwu.edu.cn} \\
  \texttt{zhangyuhe0601@nwu.edu.cn} \\
}

\begin{document}
\maketitle
\begin{abstract}
Traditionally, style has been primarily considered in terms of artistic elements such as colors, brushstrokes, and lighting. However, identical semantic subjects, like people, boats, and houses, can vary significantly across different artistic traditions, indicating that style also encompasses the underlying semantics. Therefore, in this study, we propose a zero-shot scheme for image variation with coordinated semantics. Specifically, our scheme transforms the image-to-image problem into an image-to-text-to-image problem. The image-to-text operation employs vision-language models (\textit{e.g.}, BLIP) to generate text describing the content of the input image, including the objects and their positions. Subsequently, the input style keyword is elaborated into a detailed description of this style and then merged with the content text using the reasoning capabilities of ChatGPT. Finally, the text-to-image operation utilizes a Diffusion model to generate images based on the text prompt. To enable the Diffusion model to accommodate more styles, we propose a fine-tuning strategy that injects text and style constraints into cross-attention. This ensures that the output image exhibits similar semantics in the desired style. To validate the performance of the proposed scheme, we constructed a benchmark comprising images of various styles and scenes and introduced two novel metrics. Despite its simplicity, our scheme yields highly plausible results in a zero-shot manner, particularly for generating stylized images with high-fidelity semantics.

\end{abstract}

\keywords{Image Variation \and Image Synthesis \and Style \and Style Transfer}

\section{Introduction}

\begin{figure*}[t]
\centering
\includegraphics[width=1\textwidth]{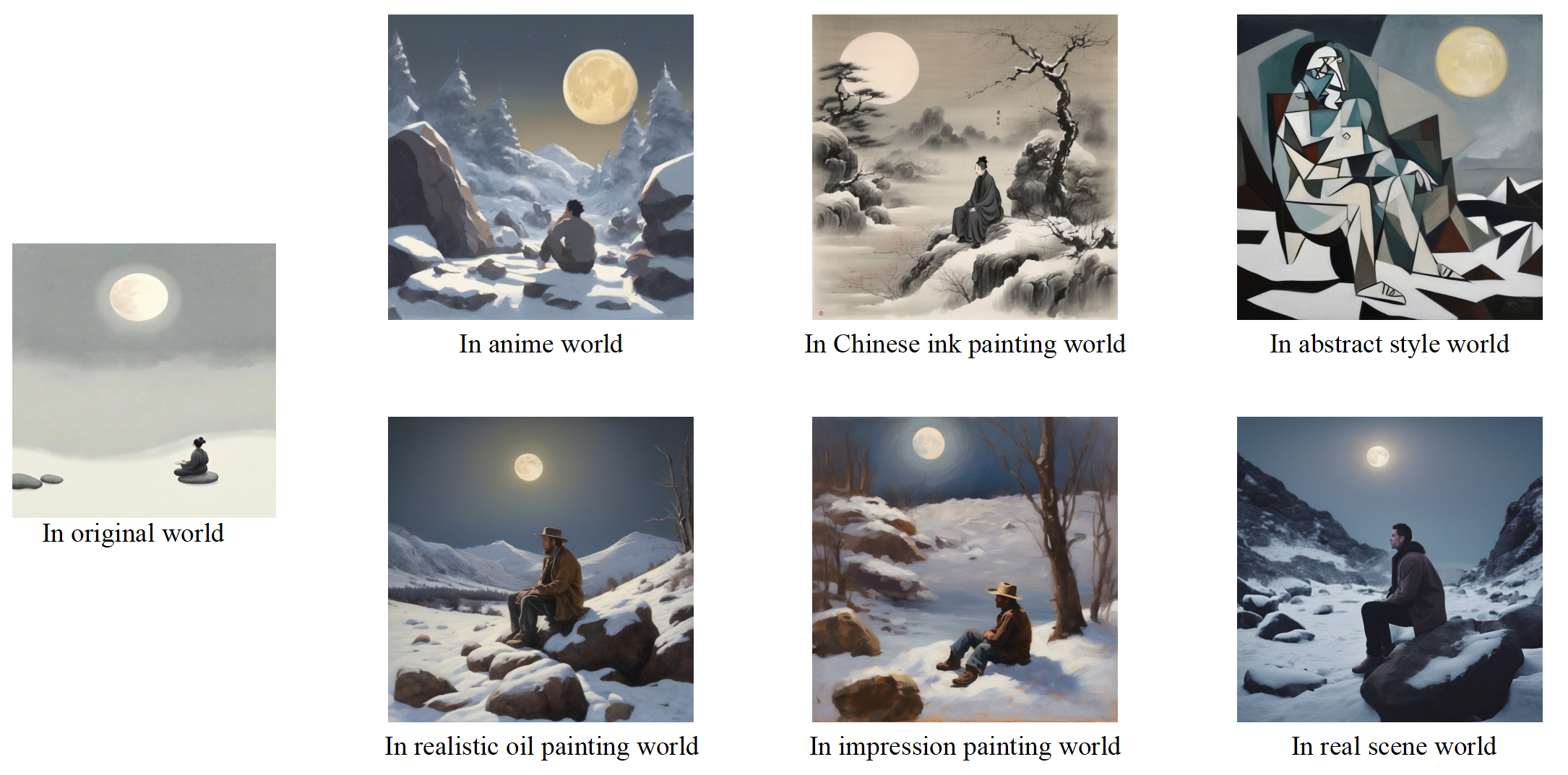} 
\caption{Given an RGB image and a style keyword, our image-to-text-to-image scheme generates the image variations in the target style with coordinated semantics. These results look like images in different worlds.}
\label{teaser}
\end{figure*}

Painting fundamentally underpins the human experience, serving as a crucial medium for expressing our hopes, dreams, fears, and emotions. Individuals from diverse cultural backgrounds employ an array of artistic methods to articulate their unique perspectives and experiences. For instance, a comparative analysis of traditional Chinese and Western art reveals significant differences in composition, form, and lighting, extending beyond mere variations in color, tone, and brushstroke. Moreover, the same semantic subjects, such as people, boats, and houses, exhibit significant variation across these diverse artistic traditions, as illustrated in Figure \ref{teaser}. Therefore, we argue that style encompasses not only artistic elements such as colors and lines but also the semantics underlying the style. Despite advancements, current style transfer approaches to varying image styles remain quite limited.

\begin{figure}[h]
\centering
\includegraphics[width=0.8\textwidth]{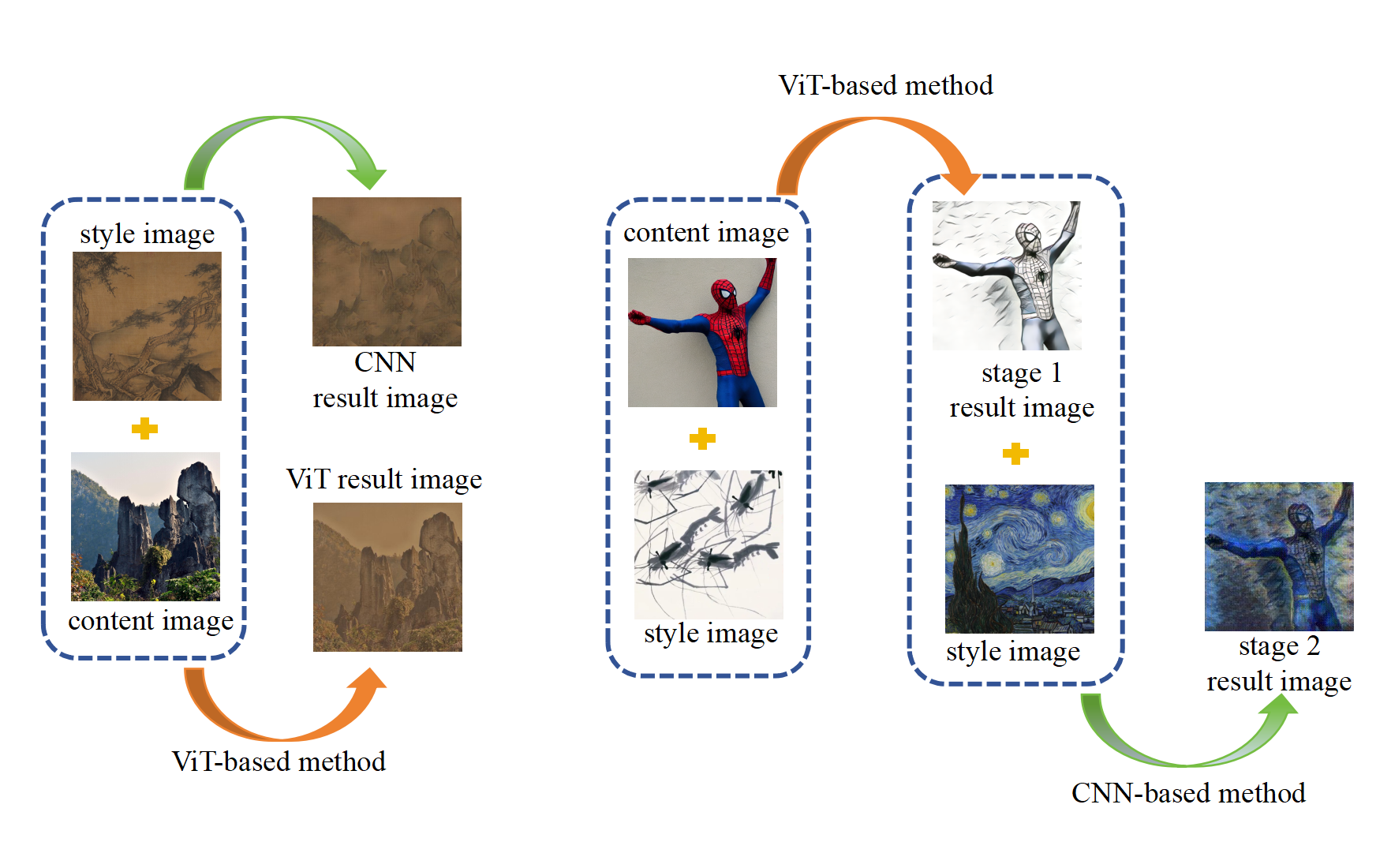}
\caption{Existing style transfer methods prioritize retaining content while adjusting color and brushstrokes, often resulting in images that lack authenticity in style. Furthermore, it is challenging to achieve satisfactory transfer results when an image with an applied style is used as the input for a continuous transfer task.}
\label{relatedwork}
\end{figure}

Existing style transfer methods, including CNN-based methods \cite{huang2017arbitrary}, GAN-based methods \cite{zhu2017unpaired} and visual Transformer-based methods \cite{li2023compact}, aim to minimize content loss, ensuring the integrity of the content. Due to the coupling of style and content in images, existing methods usually use photos as input rather than stylized images. Using distinct styles as input can cause style overlap, leading to unsatisfactory results, as shown in Figure \ref{relatedwork}. Multi-conditional image generation methods, such as SD-refiner \cite{podell2023sdxl} and Dreambooth \cite{ruiz2023dreambooth}, can enhance the original image using prompts to achieve high-quality outcomes, but their applicability is restricted to a narrow range of styles. Furthermore, both style transfer and multi-conditional generation techniques often overlook the fact that semantics can differ across styles, resulting in unfaithful representations, as illustrated in Figure \ref{relatedwork}. This shortcoming stems from the lack of datasets featuring image pairs with consistent semantics across different styles, as most existing methods rely on a supervised paradigm that depends heavily on the availability and diversity of annotated datasets. Thus, it is crucial to coordinate the semantics of the content during image style transfer to produce more faithful images.


 
Recently, several large-scale models \cite{radford2018improving,radford2019language,brown2020language,rombach2022high,podell2023sdxl} were introduced for different modalities, such as GPT3 \cite{achiam2023gpt} and Bloom for language, Stable Diffusion \cite{rombach2022high,podell2023sdxl} and DALLE-2\cite{ramesh2022hierarchical} for vision. These large-scale models are usually referred to as foundation models, and they have a broad knowledge of their domains since they were trained on large amounts of data. There are even ongoing efforts to connect these models to build a bridge between different modalities, such as Visual ChatGPT \cite{wu2023visual}, and MiniGPT-4 \cite{zhu2023minigpt}. 


Human artists first interpret the scene before creating the picture. Therefore, we propose a zero-shot learning method that transforms the image-to-image problem into an innovative image-to-text-to-image framework. By leveraging text to decouple style from content, this approach ensures both content integrity and style coherence. The image-to-text interaction focuses on extracting and describing the content and style of the image in greater detail, leading to more distinct styles. Meanwhile, the text-to-image process generates images with the same content but in various targeted styles.

Specifically, we divide the task into three parts.
First, the image is converted into natural language using the visual answering model BLIP \cite{li2022blip,li2023blip} to generate text descriptions and identify the position of each object, which can be done through BLIP-VQA \cite{li2022blip,li2023blip}. Second, the input text is interpreted using ChatGPT \cite{brown2020language, wu2023visual} to extract stylized keywords. These stylized and positional keywords are then adjusted and combined with the text descriptions to form a new prompt. Finally, the text prompts are fed into the diffusion model to redraw an image. Our approach requires only an image of any style and the text of the desired style to transform it, resulting in a new image that is high-quality, stylistically accurate, and semantically similar to the original content. Additionally, we have fine-tuned the Stable-Diffusion-xl-base \cite{podell2023sdxl} by integrating cross-attention mechanisms to improve its ability to handle a broader range of styles, including Chinese ink painting, Chinese freehand, and abstract styles.




We apply our approach to generating images in seven distinct styles: 'realistic oil painting,' 'anime,' 'Chinese ink painting,' 'impressionist oil painting,' 'abstract painting,' 'freehand Chinese painting,' and 'photographs of real scenes.' Crucially, across all these styles, we maintain content fidelity and coordinate semantics effectively.

To the best of our knowledge, our technique is the first to address the challenging problem of style-specific image variations with coordinated semantics. To evaluate this novel task, we have created a new dataset featuring these seven styles and proposed two novel evaluation metrics that assess both content fidelity and style quality of the generated images. We highlight the contributions of our method by comparing it with existing image-driven style transfer methods, text-driven style transfer methods, and multi-conditional image generation methods. Additionally, we conducted a user study to assess the content fidelity and authenticity of the synthesized images compared to alternative approaches.

\indent The main contributions can be summarized as follows:
\begin{itemize}
\item A zero-shot learning scheme for stylised image variation with coordinated semantics. 

\item Two novel metrics: weighted style mean and content matching for validating complex style transfer results are introduced.

\item A novel benchmark named \textbf{Z}ero-\textbf{s}hot \textbf{S}tyle \textbf{T}ransfer validaton Dataset (ZsSTD) containing image groups with semantic annotation in different styles, is built. ZsSTD can be used for evaluating style transfer tasks, Text-to-image tasks, and multi-conditional image generation tasks, \textit{etc}.
\end{itemize}

\section{Related Work}
\subsection{Deep Learning-Based Style Transfer}
Style transfer creates an image in a desired style by applying style features (\textit{e.g.}, texture, color, lines) to a source image. \cite{gatys2016image} pioneered the use of deep learning for style transfer with VGG-16. Later, Context-Encoder \cite{pathak2016context} introduced GANs for inpainting, followed by Pix2Pix \cite{isola2017image}, a GAN-based style transfer network that concurrently performs style transfer but requires paired data for training. It achieves style transfer by minimizing the difference between the content image and the style image while maximizing the similarity between the generated image and the style image \cite{gatys2016image}.
To solve the data problem, CycleGAN \cite{zhu2017unpaired} for unsupervised image translation is proposed. CycleGAN can learn the mapping between two domains using only unpaired samples from each domain, making CycleGAN highly versatile and applicable to a wide range of real-world problems where paired data may be difficult or impossible to obtain. Additionally, CNN-based methods often suffer from the loss of global image semantics. Recently, styTR \cite{deng2022stytr2} addressed this issue by incorporating Transformers into style transfer.

With the introduction of the text-vision model CLIP, text-driven style transfer methods have proliferated \cite{gal2022stylegan,kwon2022clipstyler}. These approaches typically leverage CLIP's ability to map relevant style features and integrate them during the decoding phase. However, reliance on a simple decoding architecture and limited textual prompts often results in suboptimal outcomes. More recently, Diffusion models have emerged as exceptional performers in generative tasks \cite{yang2023zero}, leading to the fusion of CLIP with Diffusion to create a novel paradigm for creative generation through its intricate denoising process.

\subsection{Large-scale Vision Models}

Image synthesis has rapidly advanced with the development of diffusion and large-scale models. Initially, DDPM \cite{ho2020denoising} was introduced for text-to-image tasks. Later, DDIM \cite{song2020denoising} reduced sample size from thousands to tens by predicting results after multiple denoising steps in a single iteration. DALL-E 2 \cite{rombach2022high} improved image synthesis by combining diffusion with a pre-trained CLIP model \cite{radford2021learning}, producing realistic, high-resolution images from natural language descriptions. However, state-of-the-art models like Imagen \cite{saharia2022photorealistic} require substantial hardware and extensive training time. A potential solution is to use feature compression to reduce resource dependency. The Latent Diffusion Model (LDM) \cite{ramesh2022hierarchical} shifts the diffusion process from the original image pixel space to the latent space, where the probability distribution can be obtained through trained models like \cite{kingma2013auto} or \cite{van2017neural}. This approach improves generation efficiency and reduces reliance on computing resources.


\subsection{Large-Scale Language Models} 

Large-scale models are deep neural networks with billions of parameters, trained on vast data. The Transformer architecture \cite{vaswani2017attention, xiong2020layer} initiated the era of large-scale language models, leading to developments like BERT \cite{devlin2018bert} and GPT-1 \cite{radford2018improving}. Subsequent models GPT-2 \cite{radford2019language} and GPT-3 \cite{brown2020language} expanded parameters to hundreds of billions. ChatGPT \cite{brown2020language} marked a breakthrough in open-domain Q\&A and natural language generation, with GPT-4 \cite{achiam2023gpt} further advancing the field through multi-modal understanding, dialogue memory, and advanced reasoning. ChatGPT performs sampleless learning through In-Context Learning, where a small number of labeled instances are spliced into the samples to be analyzed, which are then fed into a language model, which is used to understand the task based on the instances and give correct results. It has exhibited very strong capabilities in test datasets including TriviaQA\cite{joshi2017triviaqa}, WebQS, CoQA\cite{reddy2019coqa}, etc., even surpassing previous supervised methods in some tasks.

\section{Method}

\begin{figure*}[t]
\centering
\includegraphics[width=1\textwidth]{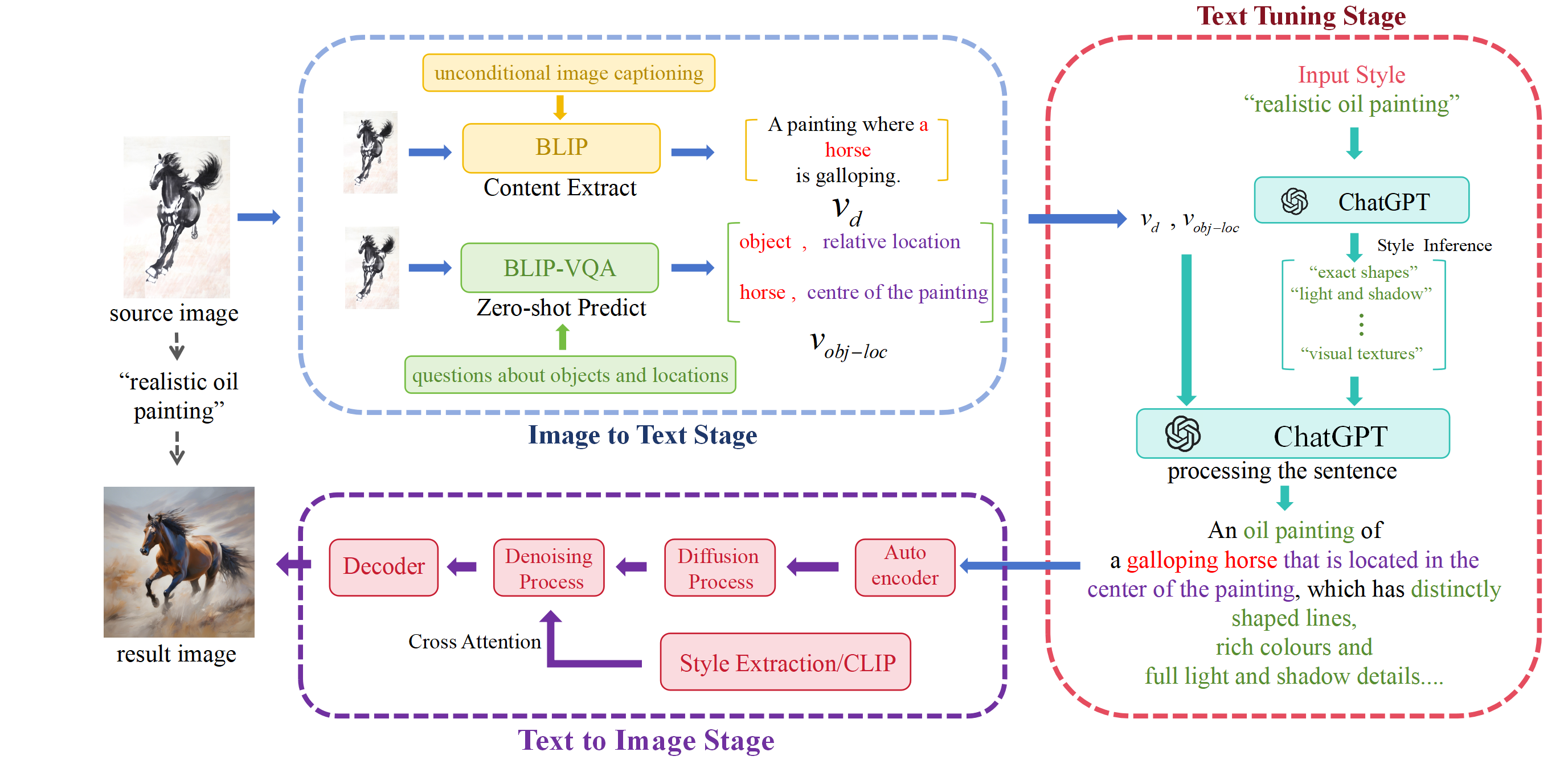} 
\caption{ Our image-to-text-to-image scheme: The source image is first input into the image-to-text module, which consists of BLIP and BLIP-VQA, to obtain the image content with the location of objects. Style keywords are then integrated with this content using ChatGPT to create a text prompt for the text-to-image module. Finally, the text-to-image module incorporating a latent diffusion model, generates the image with the same content in the desired style.}
\label{pipeline}
\end{figure*}

\begin{figure*}[t]
\centering
\includegraphics[width=1\textwidth]{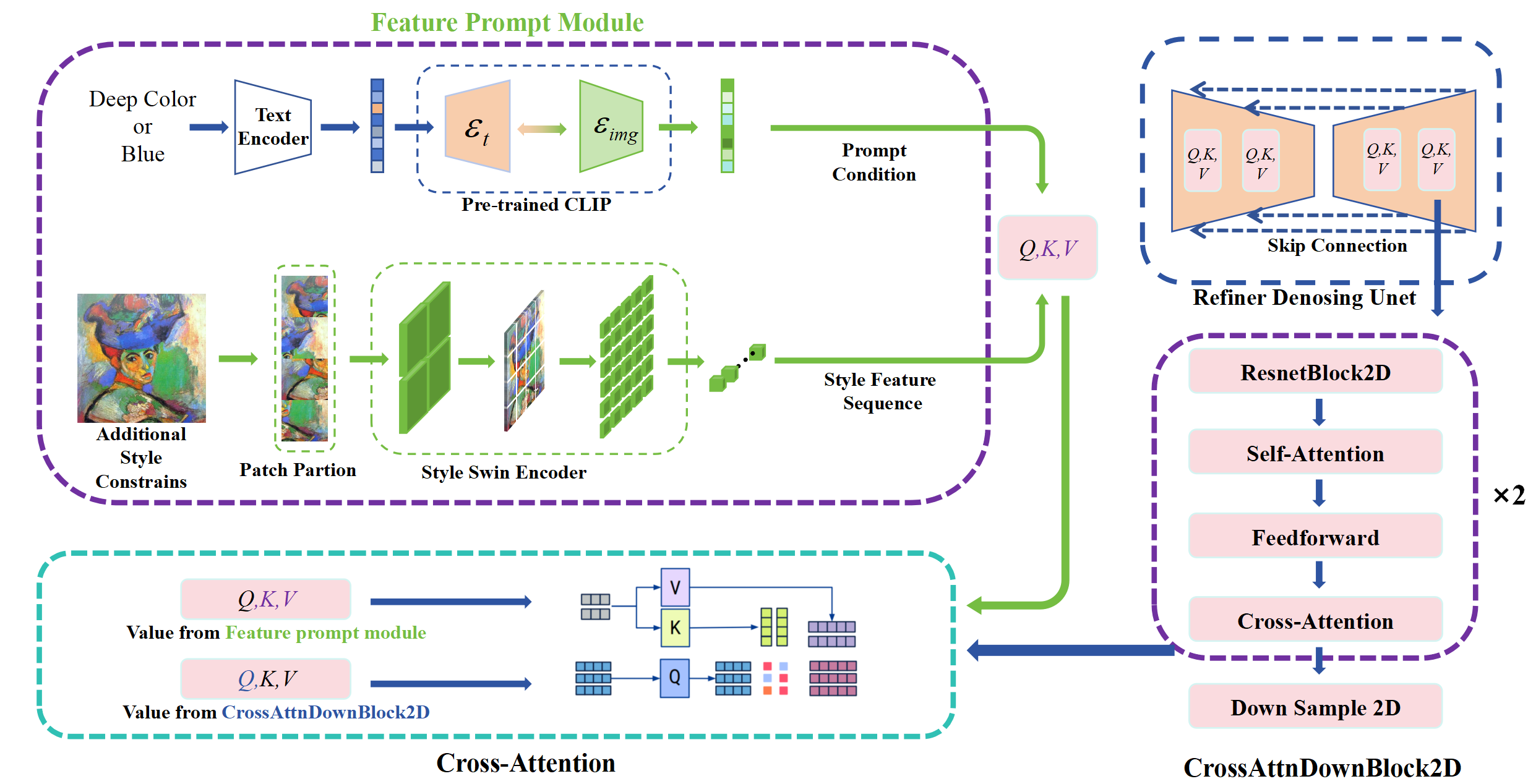} 
\caption{The Flowchart of Conditional Constraints. The feature values from CLIP or Swin style encoder are encoded to get feature sequence, which are put into CrossAttnDownBlock's Cross-attention layer subject to calculate. }
\label{conditionConstraints}
\end{figure*}

\subsection{Overview}
The pipeline of our zero-shot style transfer scheme, illustrated in Figure \ref{pipeline}, processes an image of any style—such as real scenes, oil paintings, or ink paintings, \textit{etc.}—and a style keyword. The scheme comprises three modules: an image-to-text module for extracting the image content; a text-tuning module for creating a prompt that integrates image content with style prompts, and a text-to-image module for generating an image that combines the source image's content with the specified style, rather than merely modifying colors and lines. Notably, our zero-shot approach does not require paired samples for training and is not limited to specific styles, although abstract styles are excluded due to their complexity, which makes interpretation challenging for both humans and computers.

\subsection{Image-to-Text Module}

Initially, we need to extract the content of the source image and describe it using text. This approach allows for the decoupling of content and style, thereby avoiding any influence from the source style. To address this, we propose employing a language-vision foundation model, such as BLIP2 \cite{li2023blip} and BLIP-large \cite{li2022blip}, which leverage the generalization capabilities of Large-Scale Language Models (LLMs) to reason about 2D images. In this work, the source image is input into BLIP-large \cite{li2022blip} to obtain a text vector describing the image content. Subsequently, we input the source image into BLIP-VQA \cite{li2023blip} for conditional questioning, which queries the object and its position within the image.

Consequently, this stage is formulated as an image-to-text problem.

the content of the image needs to be extracted, and we have found that text is more suitable for describing the content of the image than the image itself. Text can decouple style and content in an image to avoid other styles in the content image influencing the results. Natural language does not have the redundant information of an image than an image, but retains the description of objects and their actions. Another advantage of using natural language to describe images is that natural language clearly separates the style and the content from the image. Thus, the problem at this stage is specified as an image-to-text problem. 

Although BLIP and BLIP-VQA have very powerful inference and generalization capabilities, there are sometimes problems with recognition errors and inaccurate kinds. In order to ensure the accuracy of the object types, we use CLIP \cite{radford2021learning} to perform zero-shot predictions of the objects in the vectors. Both the image and the objects are fed into CLIP to obtain a sequence of predictions, and when the prediction value is relatively high, the recognition is considered to be accurate. If the recognition fails, the image is fed into BLIP again for secondary image generation of text, and additional conditions are used instead of unconditional.

This module outputs two text vectors: one describing the content of the image and the other detailing the objects and their positions within it.\\


\subsection{Text Tuning Module}


Given the text vectors containing the content of the image and the objects with their positions, this module serves two functions. Firstly, it translates the style into a specific description, thereby enhancing the representation of the style. For example, as illustrated in Figure \ref{pipeline}, if the input style keyword is "realistic oil painting", the detailed features corresponding to it would include rich colors, a background filled with objects, and so on. Secondly, it fuses all the provided keywords into a coherent sentence. 

This process requires extensive knowledge of art and an understanding of in-context semantics. In-context learning is the process by which a model understands a particular task and provides an adequate response to the required task \cite{abdelreheem2023zero}. LLMs are indeed proficient in-context learners, allowing them to perform well on a wide range of tasks without explicit fine-tuning.

Therefore, we integrate ChatGPT \cite{brown2020language} into this module to achieve the two desired functions. This approach involves providing the model with a few (input, expected output) pairs as examples within the input prompt when task-solving \cite{abdelreheem2023zero}. The model then generates a detailed description of the desired style. Following this, the content of the image, including the objects and their positions, along with the detailed description of the desired style, are combined into a single sentence. This sentence is then used as input for the text-to-image module. then materialize this information into a sentence. For example: we enter realistic oil painting in the input stage, we need to get the detailed features corresponding to realistic oil painting in this stage, e.g. rich colours of the oil painting, filling the background with objects, etc. 

The style text is processed by ChatGPT to get the style keyword vector. At the same time, we input the output of the previous stage, two vectors, with the keyword vector into ChatGPT for fusion, by utilising the powerful in-context understanding capabilities and text generation capabilities of ChatGPT. Subsequently a content text with style details description is obtained. This result will be used in the next stage of the text to image task. The idea is when asking the model to solve a task given a certain input, we include a few (input, expected output) pairs as examples in the input prompt\cite{abdelreheem2023zero}. The stylized text is processed by ChatGPT\cite{brown2020language} to obtain a vector of stylized keywords. At the same time, we fused the output of the previous stage, the two vectors, with the keyword vectors input to the ChatGPT species, by exploiting the powerful in-context understanding and text generation capabilities of ChatGPT.

\subsection{Text-to-Image Module}
Given the text prompt, this module is responsible for generating the image based on it. Among the many open-source models, the stable diffusion family of models exhibits excellent performance. Therefore, we use Stable-Diffusion-XL-base \cite{podell2023sdxl} to generate high-quality images. 

The Stable-Diffusion-XL-base model effectively handles many common image generation styles, such as realistic oil paintings and anime. However, it struggles with specific styles like Chinese ink painting and abstract art. Additionally, the style keywords provided by GPT-4 \cite{achiam2023gpt} sometimes overlook low-level style features, such as colors and lines, which is also a challenge for the Stable-Diffusion during generation. To overcome these limitations, we fine-tuned the Stable-Diffusion-XL-base model by integrating additional constraints in a cross-attention mechanism, the pipeline is illustrated in Figure \ref{conditionConstraints}.

In particular, our constraints are classified into text and image constraints. For text constraints, we use a pre-trained CLIP \cite{radford2021learning} to encode prompts to obtain corresponding embeddings. For single-image style constraints, we use Swin Transformer \cite{liu2021swin} to extract style embeddings. Unlike the traditional Swin Transformer, our approach focuses on style features without considering the correlation between each window and the features of other windows. Consequently, we eliminate the complex mask operation and window shifting operation, reducing both computational effort and model complexity. Instead, we use continuous window attention to extract better style features. 

The feature sequences obtained from either CLIP or the Swin Transformer are introduced into the generation process using Cross Attention in the denoising U-net. In the T-GATE study \cite{zhang2024cross}, it was found that after a few inference steps, the output of the cross-attention converges to a fixed point, typically within 5 to 10 steps. Therefore, multi-conditional constraints can be applied in subsequent steps to achieve optimal results. To ensure this, we froze the first 30 inference steps of the pre-trained Stable Diffusion-base-XL \cite{podell2023sdxl} and fine-tuned it with new prompts during the last 20 steps \cite{zhang2024cross}.

\section{Experiments}
In this section, we present the results of our approach and compare them with recently proposed state-of-the-art methods. Additionally, we evaluate the proposed method through quantitative and qualitative analysis, complemented by a user study. 

\textbf{Our goal is to capture and recreate stylized variations of an image while preserving the content semantics and coordinating the underlying style semantics, thereby enhancing the faithfulness of the output images.} Therefore, for quantitative evaluation, we introduce the \textbf{Z}ero-\textbf{s}hot \textbf{S}tyle \textbf{T}ransfer validation \textbf{D}ataset (ZsSTD) and propose two evaluation metrics to assess both content fidelity and style quality.

\begin{figure*}[t]
\centering
\includegraphics[width=1\textwidth]{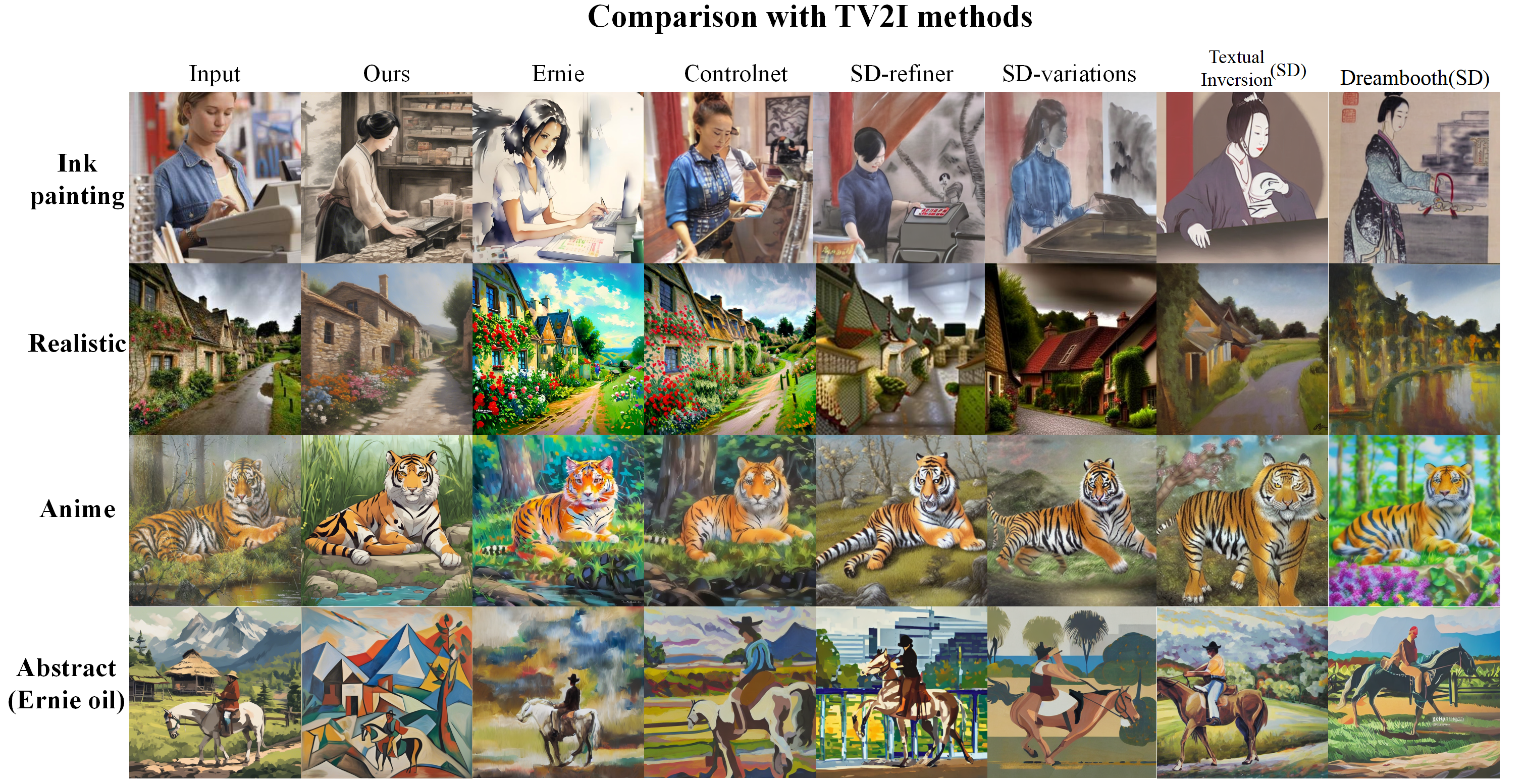} 
\caption{The comparison of visual results with multi-conditional image generation methods. Our approach more accurately preserves semantics and generates images with distinctly desired styles.}
\label{comparison1}
\end{figure*}

\subsection{Dataset, Metrics and Baselines}
\subsubsection{ZsSTD.} 
Our goal differs from traditional style transfer tasks, even though the output image is also style-specific. Therefore, to validate our approach, we collected a dataset of 5,000 images encompassing various styles of artworks from WikiArt\cite{phillips2011wiki} and the web. Specifically, the ZsSTD dataset contains seven distinct styles: 'realist oil painting,' 'anime,' 'Chinese ink painting,' 'impression oil painting,' 'abstract paintings,' 'Chinese freehand painting,' and 'real photographs.' Each style includes thousands of images with corresponding text descriptions, which are annotated to ensure accurate representation, rather than relying on existing image-to-text methods.

Some of these styles are divided into three types of scenes: human, non-human scenes, animals and plants. A large collection as well as manual labelling of the corresponding descriptions is time-consuming and labor-intensive. However, it was very helpful for us to verify the validity of the method, and it set a dataset that can be used for validation and reference for later work.

Existing style transfer methods need to be individually trained for this style in the context of a specific style transfer task. This leads to the weak generalization ability of the previous methods. Moreover, there is a lack of style-specific data for training. A model that needs to perform generalized style transfer requires a wide variety of data for training or testing. However, no dataset contains all kinds of styles. Our zero-shot style transfer method needs to verify its generalization performance and requires accurate natural language descriptions of the source images when verifying whether the semantics of the source images are the same as the result images.

\subsubsection{Metrics.} Since the images generated by our task are not tied to the style of a single image, traditional style loss \cite{gatys2016image} is insufficient for describing the styles of our generated images. Therefore, we propose a novel metric, \textit{i.e.}, \textbf{Style Mean Loss (SML)} to assess the style consistency. The SML is calculated as follows:

\begin{equation}
SML=\frac{1}{N}\sum_{i=1}^{N}\left\|Gram_{res}-Gram^i_{target}\right\|^2
\end{equation}
\noindent where $Gram^i_{target}$ is the Gram matrix of the \textit{i}-th image, ${i \in [1,2,...,N]}$, $N$ is the total number of images of this style in the dataset.


Although our output image preserves the content semantics of the input image, the actual objects are altered, rendering content loss \cite{gatys2016image} ineffective. To address this, we introduce a novel metric called the \textbf{Content Matching Score (CMS)} to measure the fidelity of content semantics. Specifically, we use the text-vision model GPT-4 \cite{achiam2023gpt} to generate textual descriptions of the images, rather than BLIP \cite{li2022blip}, as our method already incorporates BLIP, which could bias the results and create an unfair comparison with other models. The two textual descriptions are then encoded as word embeddings using a transformer \cite{vaswani2017attention}, and their cosine similarity is calculated to derive the CMS. The calculation is as follows:
\begin{equation}
CMS = \frac{\sum_{i=1}^nA_i\times B_i}{\sqrt{\sum_{i=1}^n\left(A_i\right)^2}\times\sqrt{\sum_{i=1}^n\left(B_i\right)^2}}
\end{equation}
\noindent where $A_{i}$ and $B_{i}$ represent the word vectors of the source image content embedding and the resulting image content embedding, respectively.


In addition, we employed the \textbf{Fréchet Inception Distance (FID)} to evaluate the quality of the generated images. Since our task involves text-driven style repainting, we also compute the \textbf{CLIP\_score (CLIPS)} by comparing the input style text with the output image to assess the alignment between the style and the text.

These metrics offer nuanced insights: SML quantifies the fidelity of low-level stylistic features; CLIPS assesses the overall accuracy and coherence of the image's stylistic portrayal; CMS verifies the preservation of source image content in the outputs; and FID gauges the overall visual quality and realism of the generated images.


\subsubsection{Baselines.} 

We compare our method to \textbf{12} baselines across three categories of tasks with similar focuses: (i) image-driven style transfer methods, including: CNN-based method Avatar \cite{sheng2018avatar}, AdaIn \cite{huang2017arbitrary}, flow-based method ArtFlow \cite{li2021multi}, and Transformer method styTR \cite{deng2022stytr2}; (ii) text-driven style transfer tasks, such as styleGAN-NADA \cite{gal2022stylegan}, styleCLIP \cite{patashnik2021styleclip}, CLIPstyler \cite{kwon2022clipstyler}, VQGAN-CLIP \cite{crowson2022vqgan}; and (iii) multi-conditional image generation methods, including Ernie \cite{sun2019ernie}, SD-XL Refiner \cite{podell2023sdxl}, ControNet \cite{zhang2023adding}, and Dreambooth \cite{ruiz2023dreambooth}.

\subsection{Qualitative Results} 
\subsubsection{Comparison to Image-driven Style-transfer methods.} 
\begin{figure*}[thb]
\centering
\includegraphics[width=1\textwidth]{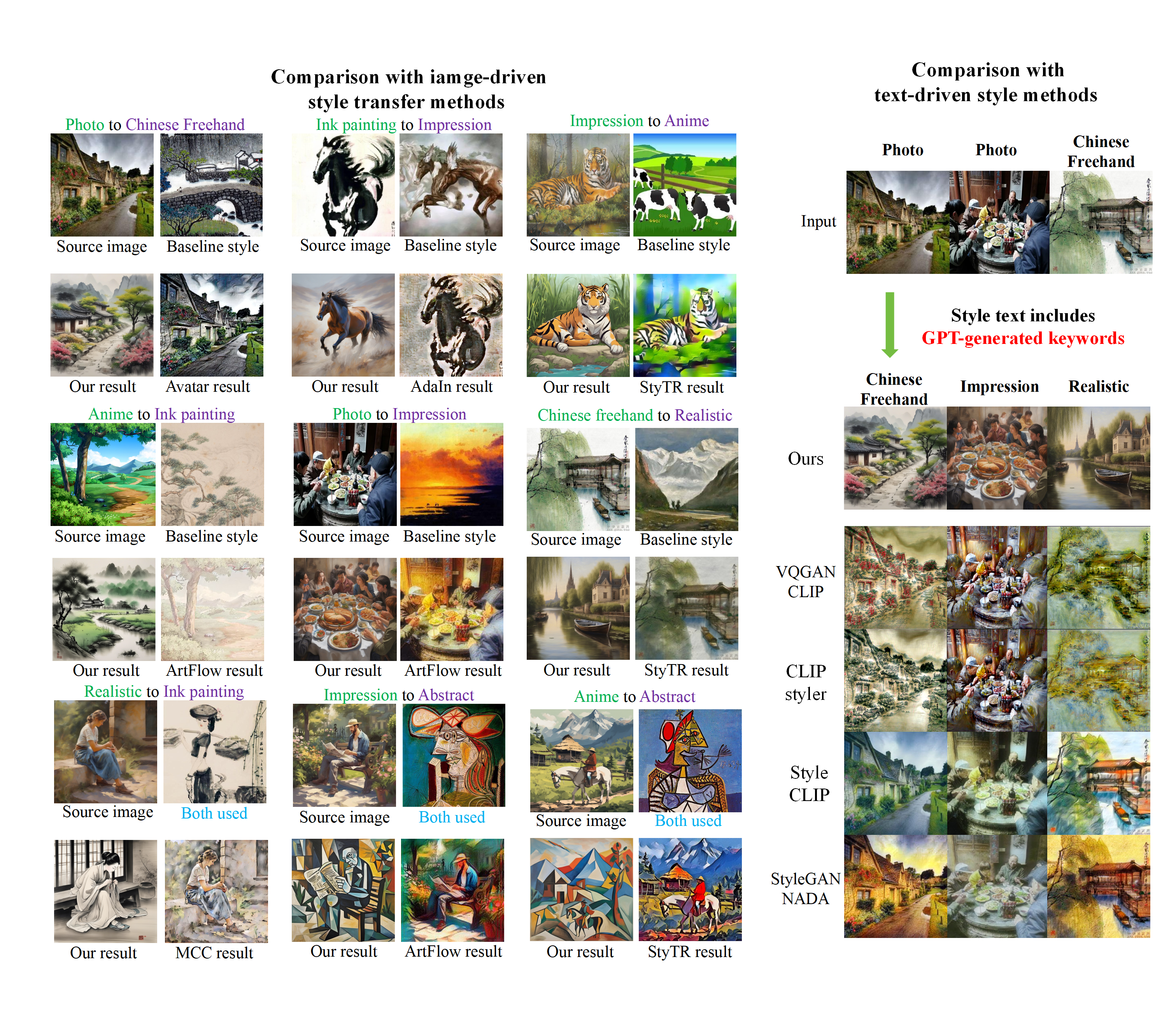}
\caption{The comparison of visual results with style transfer methods. Our method generates variations that are typically more faithful to the specific style.}
\label{comparison2}
\end{figure*}

We first conduct a comparative analysis of our method's outcomes relative to conventional style transfer approaches, which aim to preserve content and blend styles by minimizing both style loss and content loss. To ensure a thorough evaluation, we applied six distinct styles of image transformation to each category in the ZSTD dataset, generating a diverse array of stylized images for comparison, except for the style of input. The results of these various style transfers are shown in Figure \ref{comparison2}.

It can be seen that minimizing the distance between Gram matrices captures only low-level style features. When the input content image retains its original style, minimizing content loss paradoxically preserves the inherent shape and style of the input image, leading to images that exhibit merged styles. Furthermore, as shown in Figure \ref{comparison2}, for specific styles such as 'realist oil painting' and 'Chinese ink painting' or 'Chinese freehand painting', the actual objects with the same semantics, such as humans, houses, and boats, should be altered to generate more faithful images. 

\subsubsection{Comparison to Text-driven
Style-transfer methods.} Given that these text-driven style transfer baselines are based on the CLIP, we have observed that short prompts do not effectively impart style to these methods. Therefore, we also utilize GPT's reasoning capabilities to explore stylistic nuances and generate comprehensive style descriptions, ensuring a more accurate and nuanced assessment of the results. The results are presented in the Figure \ref{comparison2}. It can be observed that text-driven style transfer methods encounter similar issues to image-driven style transfer methods, such as style coupling and inconsistencies in style context. 

The experimental results conclusively demonstrate that relying solely on prompts is insufficient for achieving both style accuracy and distinction. Moreover, neither image-driven style transfer methods nor text-driven methods can fully capture the semantics underlying styles.

However, our method stands out by disentangling style from content via prompts, enriching results with stylistic nuances \& color constraints. It preserves intricate semantics, effortlessly removing source style and aligning new style seamlessly with content, excelling in complex transfers.

our groundbreaking method adeptly disentangles an image's content and style using natural language, leveraging the powerful inference capabilities of GPT. 

These keywords are then strategically paired with their networks, ensuring a more accurate and nuanced assessment of results. Some comparison results are shown in the Figure 6. 

approach adequately addresses the nuanced style context of the depicted objects or the coherent integration of image semantics, resulting in images with indistinct styles.

Upon scrutinizing the visual results, it becomes evident that both text-guided and image-based methodologies merely blend styles with content in a rudimentary fashion.

\subsubsection{Comparison to Multi-conditional Image Generation Methods.}

In Figure \ref{comparison1}, we compare the results of our method with those of multi-conditional image generation methods. The competitors often struggle with content accuracy (such as the first case) or style quality (\textit{e.g.}, the second case). Furthermore, the baseline methods encounter significant challenges when tasked with generating abstract paintings and Chinese ink paintings, primarily due to the difficulty in disentangling their distinct artistic characteristics from low-level style features. However, our method excels at capturing the content of the input image—particularly its most prominent features—and preserves sufficient detail to accurately reconstruct the object and scene. Furthermore, our method excels notably in abstract paintings and Chinese ink paintings, a testament to its proficiency in tackling the intricacies of these artistic forms.

\begin{table}[bt]
\setlength{\tabcolsep}{0.5mm}
\caption{The quantitative and user study for comparison with baselines. Best scores are in bold. Cont stands for content, Sty stands for style, and Faith stands for faithfulness.}

\label{quantCompar}
\begin{tabular}{l|cccc|ccc}
\hline

Methods & 
\multicolumn{4}{c|}{Quantitative Results} &
\multicolumn{3}{c}{User study} \\ 

& SML$\downarrow$ & CMS$\uparrow$ & FID$\downarrow$ & CLIPS$\uparrow$ 
& Cont$\uparrow$  & Sty$\uparrow$  & Faith$\uparrow$ \\
\cline{1-5} \cline{6-8}

Avatar  & 7.25 & 0.46 & 21.74 & 24.29 & 71\% & 63\% & 68\%\\

AdaIn & 7.86 & 0.33 & 19.95 & 24.42 & 77\% & 59\% & 61\% \\

Artflow & 6.95 & 0.38 & 18.45 & 23.52 & 69\% & 71\% & 73\%\\

styTR & 6.53 & 0.45 & 18.62 & 23.33 & 74\% & 73\% & 78\% \\
\hline

styleCLIP & 7.52 & 0.44 & 19.50 & 20.74 & 49\% & 55\% & 59\% \\

CLIPstyler & 8.39 & 0.45 & 23.67 & 19.48 & 58\% & 51\% & 56\% \\
\hline

TextInversion& 7.01 & 0.42 & 18.38 & 24.92 & 64\% & 71\% & 69\% \\

Dreambooth& 6.81 & 0.29 & 20.77 & 26.41 & 68\% & 82\% & 63\% \\
\hline

Ours & \textbf{6.36} & \textbf{0.57} & \textbf{17.03} & \textbf{27.42} & \textbf{81\%} & \textbf{85\%} & \textbf{87\%}\\

\hline
\end{tabular}
\end{table}

\subsection{Quantitative Results} 

In addition to qualitative assessments, we further validated the efficacy of our method using various metrics and a user study. The average performance is presented in Table \ref{quantCompar}. It can be observed that our method excels in most comparative metrics, demonstrating its superiority over the baselines in both style (as indicated by higher SML scores) and content fidelity (reflected in higher CMS and CLIPS scores). Moreover, the higher FID scores also indicate that the quality of the image is closer to the Ground Truth in the style dataset, which implies that the resulting image has better quality. \\
\indent We also conducted a questionnaire of 15 comparative questions with 75 users from different majors and professions, including 25 students from art majors, 25 teachers teaching philosophy, and 25 professionals working with computers, avoiding stereotyping the style transfer results from a certain group of people. Each question presents a set of input images along with one generated image from each method. Users are asked to rate each image on content consistency, content fidelity, style clarity, and faithfulness of the image, \textit{etc.}, with scores ranging from 0\% (strongly disagree) to 100\% (strongly agree). We then compiled the average scores, as presented in Table \ref{quantCompar}. We observe a strong preference for our method in both content fidelity and style distinction, indicating that it has received widespread recognition and acclaim from most participants. 
\section{Conclusion}

We propose a novel stylized image variation method using an image-to-text-to-image scheme within a zero-shot learning framework. This approach transforms images into specific styles while preserving content semantics and effectively decoupling content from style through natural language. To address styles like Ink painting, Chinese freehand, and abstract art, we integrated a cross-attention mechanism to fine-tune stable diffusion, ensuring robust generalization. Additionally, we introduced new datasets and metrics tailored for evaluating stylized image variations, validating our method and providing a foundation for future research.

\indent In anticipation of future tasks, we aim to employ sketches alongside text as dual constraints for content planning, thereby mitigating issues of content leakage.

\subsection{Limited and Future Works} Despite our method's commendable performance in introducing style, relying solely on natural language to preserve semantics during the transfer process remains insufficient. Furthermore, the inherent high randomness of the diffusion model during generation introduces discrepancies between the produced content and the original outcomes, posing challenges for refinement. In anticipation of future tasks, we aim to employ sketches alongside text as dual constraints for content planning, thereby mitigating issues of content leakage. Additionally, we intend to incorporate discriminators to regulate the randomness of the diffusion process, striving for optimal outcomes in variation tasks.

\section*{Acknowledgments}
This research was support by the National Natural Science Foundation of China (Grant No. 62271393),  We also would like to express our sincere gratitude to the editor and five anonymous reviewers for their valuable comments, which have greatly improved this paper.

\bibliographystyle{unsrt}  
\bibliography{references}

\end{document}